\documentclass[runningheads]{llncs}
\usepackage{acro}
\input{abbreviation.acn}
\usepackage[T1]{fontenc}
\usepackage{graphicx,verbatim}
\usepackage{booktabs}
\usepackage{multirow}
\usepackage{amsmath}
\usepackage{amssymb}
\usepackage{arydshln}
\usepackage[colorlinks]{hyperref}
\usepackage{marvosym}

\usepackage{cleveref}
\usepackage{bm}

\begin{document}
%
%\title{CT-G3D: CT/X-ray Registration via Joint Learning of Radiative Gaussian Reconstruction and 3D/3D Registration}
\title{RadGS-Reg: Registering Spine CT with Biplanar X-rays via Joint 3D Radiative Gaussians Reconstruction and 3D/3D Registration}

\titlerunning{RadGS-Reg}

%
\begin{comment}  
% Removed for anonymized MICCAI 2025 submission
\author{First Author\inst{1}\orcidID{0000-1111-2222-3333} \and
Second Author\inst{2,3}\orcidID{1111-2222-3333-4444} \and
Third Author\inst{3}\orcidID{2222--3333-4444-5555}}

\authorrunning{F. Author et al.}

\institute{Princeton University, Princeton NJ 08544, USA \and
Springer Heidelberg, Tiergartenstr. 17, 69121 Heidelberg, Germany
\email{lncs@springer.com}\\
\url{http://www.springer.com/gp/computer-science/lncs} \and
ABC Institute, Rupert-Karls-University Heidelberg, Heidelberg, Germany\\
\email{\{abc,lncs\}@uni-heidelberg.de}}

\end{comment}

\author{Ao Shen\inst{1} %1
\and Xueming Fu\inst{2} %2
\and Junfeng Jiang\inst{3} $^{\href{mailto:jiangjf.hhu@gmail.com}{\textrm{\Letter}}}$ 
\and Qiang Zeng\inst{3}
\and Ye Tang\inst{1}
\and Zhengming Chen\inst{1}
\and Luming Nong\inst{4}
\and Feng Wang\inst{5}
\and S. Kevin Zhou\inst{2} $^{\href{mailto:skevinzhou@ustc.edu.cn}{\textrm{\Letter}}}$}

 %% Added for anonymized MICCAI 2025 submission
\authorrunning{A. Shen et al.}
\institute{College of Information Science and Engineering, Hohai University (HHU), Changzhou Jiangsu, 213200, China \and
School of Biomedical Engineering, Division of Life Sciences and Medicine, University of Science and Technology of China (USTC), Hefei Anhui, 230026, China\\ \email{skevinzhou@ustc.edu.cn} \and
College of Artificial Intelligence and Automation, HHU, Changzhou Jiangsu, 213200, China\\ \email{jiangjf.hhu@gmail.com} \and
The Third Affiliated Hospital of Nanjing Medical University, Changzhou Jiangsu, 213164, China\and
Tuodao Medical Technology Co., Ltd., Nanjing Jiangsu, 210012, China  \\
}

% \email{email@anonymized.com}

\maketitle              % typeset the header of the contribution
% $^\textrm{\Letter}$ These authors contributed equally and should be considered joint corresponding authors.

%
\begin{abstract}
Computed Tomography (CT)/X-ray registration in image-guided navigation remains challenging because of its stringent requirements for high accuracy and real-time performance. Traditional "render and compare" methods, relying on iterative projection and comparison, suffer from spatial information loss and domain gap. 3D reconstruction from biplanar X-rays supplements spatial and shape information for 2D/3D registration, but current methods are limited by dense-view requirements and struggles with noisy X-rays. To address these limitations, we introduce RadGS-Reg, a novel framework for vertebral-level CT/X-ray registration through joint 3D Radiative Gaussians (RadGS) reconstruction and 3D/3D registration. Specifically, our biplanar X-rays vertebral RadGS reconstruction module explores learning-based RadGS reconstruction method with a Counterfactual Attention Learning (CAL) mechanism, focusing on vertebral regions in noisy X-rays. Additionally, a patient-specific pre-training strategy progressively adapts the RadGS-Reg from simulated to real data while simultaneously learning vertebral shape prior knowledge. Experiments on in-house datasets demonstrate the state-of-the-art performance for both tasks, surpassing existing methods. The code is available at: \href{https://github.com/shenao1995/RadGS_Reg}{github.com/shenao1995/RadGS\_Reg}.

% Authors must provide keywords and are not allowed to remove this Keyword section.

\end{abstract}
\renewcommand{\thefootnote}{}%
\footnote{$^{{\textrm{\Letter}}}$ Junfeng Jiang and S. Kevin Zhou are joint corresponding authors.}%
\addtocounter{footnote}{-1}
\section{Introduction}
% Image-guided navigation is widely considered for minimally invasive surgery and plays an important role in enabling mixed reality, as well as robot-assisted workflows~\cite{Unberath21Suvey}. A key component of image-guided navigation is rigid 2D/3D registration, which allows us to estimate the spatial relationship between patient preoperative 3D \ac{CT} scans and 2D X-rays. In practice, only a limited number of X-rays can be acquired, typically within a limited angle range, resulting in a lack of spatial information in the projection direction and in consequence in spatial ambiguities. Low-quality X-rays (e.g., motion artifacts, scatter radiation, etc.) make registration in this setting particularly challenging.

\ac{CT}/X-ray registration has been extensively studied as an unresolved issue in the realm of image-guided navigation, characterized by its stringent requirements for accuracy and real-time performance~\cite{grupp2020automatic}. Over the past two decades, numerous methods have been devised to address these critical clinical demands; however, persistent challenges remain. Primarily, akin to the "render-and-compare" paradigm in computer vision~\cite{labbe2022megapose,ponimatkin2022focal,de2022cendernet}, the majority of traditional approaches have been predominantly projection-based. These methods iteratively refine the \ac{CT} volume's pose by repeatedly projecting the 3D \ac{CT} volume onto the detector plane, followed by a comparison of the resultant projected image, known as \ac{DRR}, with the target X-ray. Nonetheless, the \ac{DRR} image inherently loses information from its 3D \ac{CT} counterpart. Furthermore, contingent upon the nature of the projection process, various strategies have been proposed to mitigate the domain gap between DRR and X-ray~\cite{zhang2023,gao2023fully}, as well as to enhance the efficiency of DRR generation~\cite{otake2011intraoperative,gopalakrishnan2024intraoperative}. The advent of the \ac{3DGS} offers the potential to significantly reduce the domain gap and improve efficiency substantially~\cite{kerbl20233d,bao20253d}. Additionally, existing methods~\cite{naik2022realistic,jaganathan2023self} for \ac{CT}/X-ray registration concerning spinal registration often fail to account for variations in the patient's spinal poses as captured in \ac{CT} and X-rays.

Inspired by recent progress in 3D reconstruction in the field of computer vision, we present a novel approach termed \textbf{RadGS-Reg} to joint learning of \textbf{Rad}iative \textbf{G}aussians~\cite{zha2024r} Reconstruction and 3D/3D \textbf{Reg}istration for CT/X-ray Registration. This methodology entails the transformation of biplanar X-rays inputs into \ac{RadGS}, which is subsequently registered with the provided \ac{CT} volume. The task of accurately reconstructing \ac{RadGS} using only biplanar X-rays presents a significant challenge. To address this, we unveil the synergistic interaction between 3D reconstruction and \ac{CT}/3D Gaussians registration. Specifically, the shape of the preoperative \ac{CT} volume constitutes the ultimate target within the 3D reconstruction process. Concurrently, the pose of the \ac{RadGS} derived from biplanar X-rays is registered with the target pose of the preoperative \ac{CT} volume. To mitigate the issue of spinal pose variation in \ac{CT} and X-ray, we utilize the visually-grounded, region-specific vertebral-level identified in X-rays, along with the vertebral-level in the \ac{CT} volume segmented using existing methods~\cite{payer2020coarse} for vertebrae segmentation. To address the complication posed by the overlap of adjacent vertebral joints, which may impede the reconstruction of individual vertebrae, we integrate \ac{CAL}~\cite{rao2021counterfactual}, concentrating on vertebra regions to enhance reconstruction accuracy. 

% Beyond \ac{CT}/X-ray registration, the resultant reconstruction of \ac{RadGS} also facilitates the potential application of Novel View Synthesis (NVS) for intraoperative navigation~\cite{cai2025radiative}.

Our primary contributions are delineated as follows. (1) A comprehensive deep learning framework for CT/X-ray registration through sequential 3D reconstruction and 3D/3D registration, leveraging the synergy between reconstruction and registration to enhance accuracy and efficiency. (2) A vertebral-level approach using integrated \ac{CAL} to reduce interference from adjacent vertebrae, focusing on vertebral regions during 3D reconstruction from X-rays. (3) Improved training efficiency through module-aware and data-aware strategies, with results surpassing state-of-the-art methods in 3D reconstruction and registration.

\section{Method}
\begin{figure}[t]
\includegraphics[width=\textwidth]{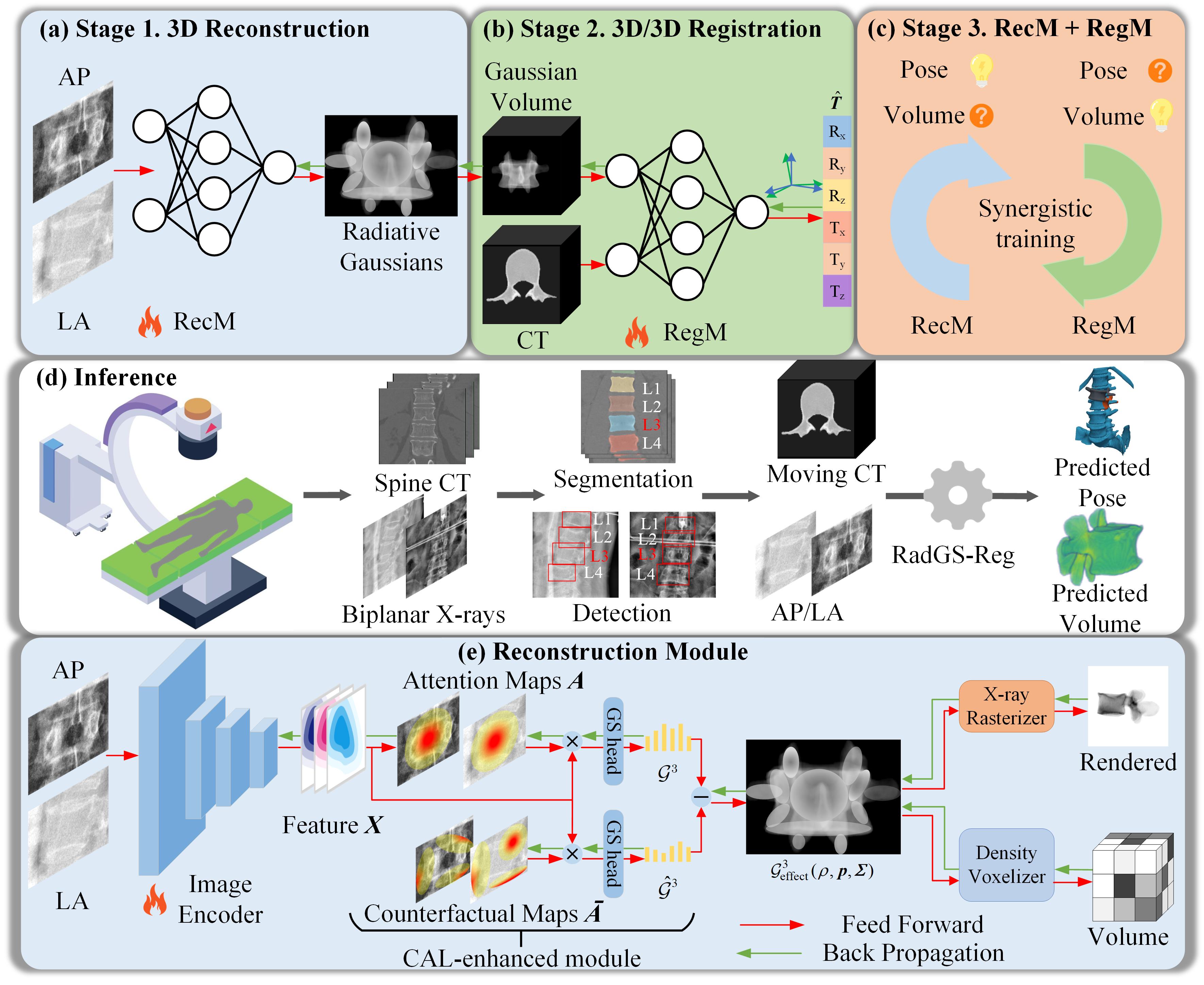}
\caption{The overall framework of RadGS-Reg (a-d) and the biplanar X-rays \ac{RadGS} reconstruction module (e). The framework consists of four parts: (a) 3D reconstruction, (b) 3D/3D registration, (c) synergistic training, and (d) inference phase. } \label{fig1}
\end{figure}

% Overview of the overall process of the RAPID method.
The proposed end-to-end deep learning framework is comprehensively depicted in Fig. \ref{fig1}, consisting of two fundamental components: the 3D reconstruction module and the 3D/3D registration module, denoted RecM and RegM, respectively. Specifically, as shown in Fig.~\ref{fig1}(a), RecM utilizes biplanar X-rays as inputs to yield predictions of \ac{RadGS} $\mathcal{G}^3= \left\{\bm{G}_i^3(\rho_i, \bm{p}_i, \bm{\Sigma}_i) \right\}_{i=1,\ldots,N}$, where $\rho_i$, $\bm{p}_i\in \mathbb{R}^3$ and $\bm{\Sigma}_i \in \mathbb{R}^{3 \times 3}$ are learnable parameters representing central density, position, and covariance, respectively. The predicted \ac{RadGS} are voxelized into a volumetric form and fed into the registration module (Fig.~\ref{fig1}(b)) along with the input \ac{CT} volume, with RegM outputting the intraoperative reconstructed volume 6-\ac{DOF} pose relative to the preoperative \ac{CT}.

% These predicted \ac{RadGS} are subsequently voxelized into a volumetric form and introduced into the registration module (Fig.~\ref{fig1}(b)), together with the input \ac{CT} volume. The RegM ultimately outputs the intraoperative reconstructed volume rigid 6-\ac{DOF} pose relative to the preoperative CT volume.

% During the inference phase (as illustrated in Fig.~\ref{fig1}(d)), the biplanar X-rays with camera parameters, and \ac{CT} volume are preprocessed by extracting the same level vertebrae using X-ray detection~\cite{jiang2024ablspinelevelcheck} and CT segmentation~\cite{payer2020coarse} methods before being fed into our trained RadGS-Reg model, enabling vertebral-level reconstruction and registration. The ultimate output comprises the reconstruction module's results, specifically the reconstructed Gaussians volume of the X-rays, alongside the output from the registration module, namely the rigid registration of the input \ac{CT} volume. Subsequently, we elucidate the procedures for training the reconstruction and registration modules, respectively.

During inference (as illustrated in Fig.~\ref{fig1}(d)), biplanar X-rays with camera parameters and \ac{CT} volume are preprocessed using X-ray detection~\cite{jiang2024ablspinelevelcheck} and CT segmentation~\cite{payer2020coarse} to extract same-level vertebrae before input to RadGS-Reg, enabling vertebral-level reconstruction and registration. The output includes both the reconstructed volume from biplanar X-rays and the rigidly registered \ac{CT} volume. The training procedures for reconstruction, registration, and their synergistic training are detailed separately.

\subsubsection{3D Reconstruction Module} Our RecM consists of four sub-modules: an image encoder, a \ac{CAL}-enhanced module, an X-ray rasterizer~\cite{zha2024r}, and a density voxelizer~\cite{zha2024r}, as illustrated in Fig.~\ref{fig1}(e). The structure of RecM is described below.

To address the challenges of blurred vertebral features in real X-rays, we propose adding a \ac{CAL} module~\cite{rao2021counterfactual} before the \ac{GSHead}, consisting of a sequence of a linear layer, a ReLU layer, and another linear layer. The \ac{CAL} module guides the network to focus on factual regions by leveraging counterfactual causality to quantify and optimize attention quality. Specifically, given input features $\bm{X}$ and factual attention maps $\bm{A}$, we leverage an intervention operation $do(\cdot)$ to generate counterfactual attention maps $\bm{\Bar{A}}$ by replacing the factual attention maps $\bm{\Bar{A}}$ with random weights. The final \ac{RadGS} prediction $\mathcal{G}^3_{\text{effect}}$ can be expressed as the difference between the factual prediction $\mathcal{G}^3$ and its counterfactual alternative $\hat{\mathcal{G}}^3$, as shown in Eq. (\ref{eq:CAL}).
\begin{equation}
\mathcal{G}^3_{\text{effect}} = \mathbb{E}_{A \sim \gamma} \left[ Y(A=\bm{A}, X=\bm{X}) - Y \left( do(A={\bm{\Bar{A}}}), X=\bm{X} \right) \right],
\label{eq:CAL}
\end{equation}
where $\gamma$ is the distribution of counterfactual attentions, $Y$ is the \ac{GSHead}.

Then the predicted \ac{RadGS} are rendered by X-ray rasterizer and concurrently voxelized by a density voxelizer. Finally, the overall reconstruction loss Eq. (\ref{eq:cal_recon}) is defined as follows.
\begin{equation}
    \mathcal{L}_{CAL-rec} = \mathcal{L}_1(\bm{I}_{effect}, \bm{I}_m) + \lambda_{\text{1}} \mathcal{L}_{SSIM}(\bm{I}_{effect}, \bm{I}_m) + \lambda_{\text{2}} \mathcal{L}_{tv}(\bm{V}_{effect}) + \mathcal{L}_{rec},
\label{eq:cal_recon}
\end{equation}
where $I_\text{effect}$ and $V_{\text{effect}}$ represent the projected image and volume, respectively, obtained by $\mathcal{G}^3_{\text{effect}}$ through the rasterizer and voxelizer, while $I_m$ denotes the measured projection. In addition to utilizing photometric L1 loss $\mathcal{L}_1$ and \ac{SSIM}~\cite{wang2004image} loss $\mathcal{L}_{SSIM}$, we further incorporate a 3D total variation (TV)~\cite{wang2008new} regularization term $\mathcal{L}_{TV}$ as a homogeneity prior for tomography. The loss $\mathcal{L}_{rec}$ also comprises $\mathcal{L}_1$, $\mathcal{L}_{SSIM}$, and $\mathcal{L}_{TV}$, but its input is derived from the output generated from the original \ac{RadGS} $\mathcal{G}^3$.

% During training, optimizing this objective Eq.\ref{eq:cal_recon} encourages the RecM to learn attention maps that are beneficial for vertebral reconstruction, thereby suppressing biased or spurious correlations (e.g., regions unrelated to vertebral morphology) and emphasizing areas causally linked to vertebral reconstruction features.

% \begin{figure}[ht]
% \includegraphics[width=\textwidth]{Figs/recM.jpg}+
% \caption{The overall framework of recM.} \label{fig2}
% \end{figure}

\subsubsection{3D/3D Registration Module} The RegM processes the concatenated Gaussian volume channel-wise and the preoperative CT volume $V_{rec}$ as input, and regresses the reconstructed volume rigid 6-\ac{DOF} pose relative to the CT volume. We simply employ an image encoder and a pose head for pose regression. The registration loss is defined as follows.

\begin{equation}
    \mathcal{L}_{reg} =\mathcal{L}_{NCC}(\bm{V}_{CT}\cdot\bm{\hat{T}} , \bm{V}_{rec}) + \mathcal{L}_{SSIM}(\bm{V}_{CT}\cdot\bm{\hat{T}} , \bm{V}_{rec}) + \lambda \mathcal{L}_{geo} (\bm{T}_{gt}, \bm{\hat{T}}),
\label{eq:reg_loss}
\end{equation}
where $\mathcal{L}_{NCC}$ represents a 3D-extended of the normalized cross-correlation~\cite{penney1998comparison} loss, and $\mathcal{L}_{geo}$ is utilized to compute the geodesic distance~\cite{salehi2018real} between the estimated $\bm{\hat{T}}$ and \ac{GT} poses $\bm{T}_{gt}$. 

\subsubsection{RadGS-Reg Training Strategy} We implement module-aware and data-aware strategies to train RadGS-Reg model. For module-aware training, we first train the 3D reconstruction (Fig.~\ref{fig1}(a)) and registration (Fig.~\ref{fig1}(b)) modules independently, followed by synergistic training (Fig.~\ref{fig1}(c)) with the total loss, as shown in Eq. (\ref{eq:tota_loss}):
\begin{equation}
    \mathcal{L}_{total} = \mathcal{L}_{reg}+ \mathcal{L}_{CAL-rec},
\label{eq:tota_loss}
\end{equation}
making the RadGS-Reg model-agnostic. For data-aware training, we adopt a three stage pre-training strategy, as detailed in Sec. \ref{sec:exp_detail}.

% In particular, we implement both module-aware and data-aware strategies to facilitate the training process of our end-to-end framework. In the context of module-aware training, we independently train the standalone 3D reconstruction module (as illustrated in Fig.~\ref{fig1}(a)) and the standalone registration module (depicted in Fig.~\ref{fig1}(b)). Subsequently, these modules undergo synergistic training (as shown in Fig.~\ref{fig1}(c)). The total training loss is formulated as shown in Eq. (\ref{eq:tota_loss}):
% \begin{equation}
%     \mathcal{L}_{total} = \mathcal{L}_{reg}+ \mathcal{L}_{CAL-rec},
% \label{eq:tota_loss}
% \end{equation}
% This synergistic training makes the Rad-GS framework model-agnostic. Regarding data-aware training, we employ a three stage pre-training strategy, as detailed in Sec. \ref{sec:exp_detail}.

\section{Experiments}
\subsection{Data Preparation}
Our datasets are sourced from the publicly available dataset VERSE '20~\cite{VERSE20} and an in-house dataset collected from a medical institution. 

The VERSE '20 dataset includes 253 lumbar spine \ac{CT} scans extracted from 300 cases, containing a total of 1,280 vertebrae. \ac{DRR}s were synthesized using a C-arm simulator (976×976 resolution, 1,124 mm focal length) with ±15° perturbations in \ac{AP} and \ac{LA} views, yielding 32,000 biplanar \ac{DRR} pairs for RecM first stage pre-training. For RegM, 1,280 \ac{CT} pairs were generated through pose sampling with parameters randomly sampled: rotations in [-20°, 20°] and translations in [-50 mm, 50 mm].

The in-house dataset includes 10 real-scent intraoperative cases, each containing preoperative CT volumes (512×512×401 voxels, spacing: 0.3×0.3×0.625 mm) and paired AP/LA X-rays (976×976 resolution, 1,124 mm focal length). Thirty vertebral-level samples were extracted and utilized for the second and third stage of pre-training with five-fold cross-validation. \ac{GT} poses (CT-to-C-arm coordinate transformation) were recorded during C-arm circular acquisitions to ensure geometric consistency.

\subsection{Experimental Setup and Evaluation Metrics}
\label{sec:exp_detail}
To implement the data-aware strategy, we proposed a three stage pre-training strategy for the RadGS-Reg, training on three distinct datasets: publicly available \ac{CT} volumes with DRRs, our proprietary real dataset, and the target \ac{CT} volume with DRRs. First, the model was pre-trained on the VERSE '20 dataset using simulated data, followed by pre-training on the in-house dataset with five-fold cross-validation. For each fold, the validation set of preoperative \ac{CT} volumes generated 1,800 biplanar DRR pairs and 600 CT pairs for the third stage. Results were analyzed using five-fold cross-validation on real X-ray data from the in-house dataset.

The evaluation results are presented in two main aspects: first, the \ac{SSIM} and \ac{PSNR} were used to gauge the quality of the reconstructed vertebra volumes. Second, the \ac{mTRE}, \ac{CR}, time efficiency, and \ac{SR} were used to measure the performance of 2D/3D registration~\cite{zhang2023}. The success of \ac{SR} was defined as the \ac{mTRE} of less than 2 mm, and the acceptable \ac{CR} interval was set to 5 mm.

\begin{table}[ht]
\caption{Our method was compared with the existing approaches in both reconstruction and registration. Statistical significance was evaluated against each method, with an asterisk ($*$) indicating $p$ < 0.05 (mean±std, best results in bold).}
\label{Compared}
\centering
\begin{tabular}{l|c|cc|cccc}
    \hline
    \multirow{2}{*}{Method} & \multirow{2}{*}{*} & \multicolumn{2}{c|}{Reconstruction} & \multicolumn{4}{c}{Registration} \\ 
    \cline{3-8}
    & & SSIM(\%) $\uparrow$ & PSNR(dB) $\uparrow$ & \ac{mTRE}(mm)$\downarrow$ & \ac{CR}(mm)$\uparrow$ & \ac{SR}(\%)$\uparrow$ & RT(s)$\downarrow$ \\
    \hline
    DiffVox & \checkmark & 15.48±11.42 &10.22±0.76 &- &- &- &- \\
    SAX-NeRF & \checkmark & 57.23±8.96 & 16.17±0.55 &- &- &- &-\\
    3DGR & \checkmark & 39.93±12.35 &15.06±2.35 &- &- &- &- \\
    R$^2$-GS & \checkmark & 52.36±2.21 & 15.79±0.54 &- &- &- &- \\
    \hdashline
    DiffPose & \checkmark &- &- & 10.49±9.90 & 10-15 & 20.00 & 139.24 \\
    DDGS-CT & \checkmark &- &- & 13.90±11.84 & 15-20 & 16.67 & 72.99 \\
    TS-SAR & \checkmark &- &- & 4.89±1.20 & 0-5 & 6.67 & 1.72 \\
    Ours &- & \textbf{94.51±0.39} & \textbf{28.80±0.33} & \textbf{1.14±1.01} & \textbf{20-25} & \textbf{93.33} & \textbf{0.82} \\
    \hline
\end{tabular}
\end{table}

% All experiments were performed on an AMD Core EYPC 7R32 48-core processor with 128 GB RAM and an NVIDIA GeForce RTX 4090 GPU with 24 GB memory. During training, the Adam optimizer was used to minimize the loss function. The RadGS-Reg was trained for 300 epochs with an initial learning rate of 0.1, which decays by 0.5 every 50 epochs. The batch size was set to 8 to accommodate GPU memory. Lastly, following the recommendations from prior studies, we set $\lambda_{1}=0.2$ and $\lambda_{2}=0.05$ in Eq. (\ref{eq:cal_recon}) as per \cite{zha2024r}, and $\lambda=0.02$ in Eq. (\ref{eq:reg_loss}) as suggested by \cite{gopalakrishnan2024intraoperative}.

All experiments were conducted on an AMD Core EYPC 7R32 48-core processor with 128 GB RAM and an NVIDIA GeForce RTX 4090 GPU (24 GB memory). The Adam optimizer was used for training, which lasted 300 epochs with an initial learning rate of 0.1, decaying by 0.5 every 50 epochs. The batch size was set to 8 to fit within GPU memory. For hyperparameters, we used $\lambda_{1}=0.2$ and $\lambda_{2}=0.05$ in Eq. (\ref{eq:cal_recon}) as per \cite{zha2024r}, and $\lambda=0.02$ in Eq. (\ref{eq:reg_loss}) as suggested by \cite{gopalakrishnan2024intraoperative}.

% \subsubsection{Implementation Details}
% \subsubsection{Evaluation Metrics}

\subsection{Results}
\subsubsection{Comparison with Existing Methods}
\begin{figure}[ht]
\includegraphics[width=\textwidth]{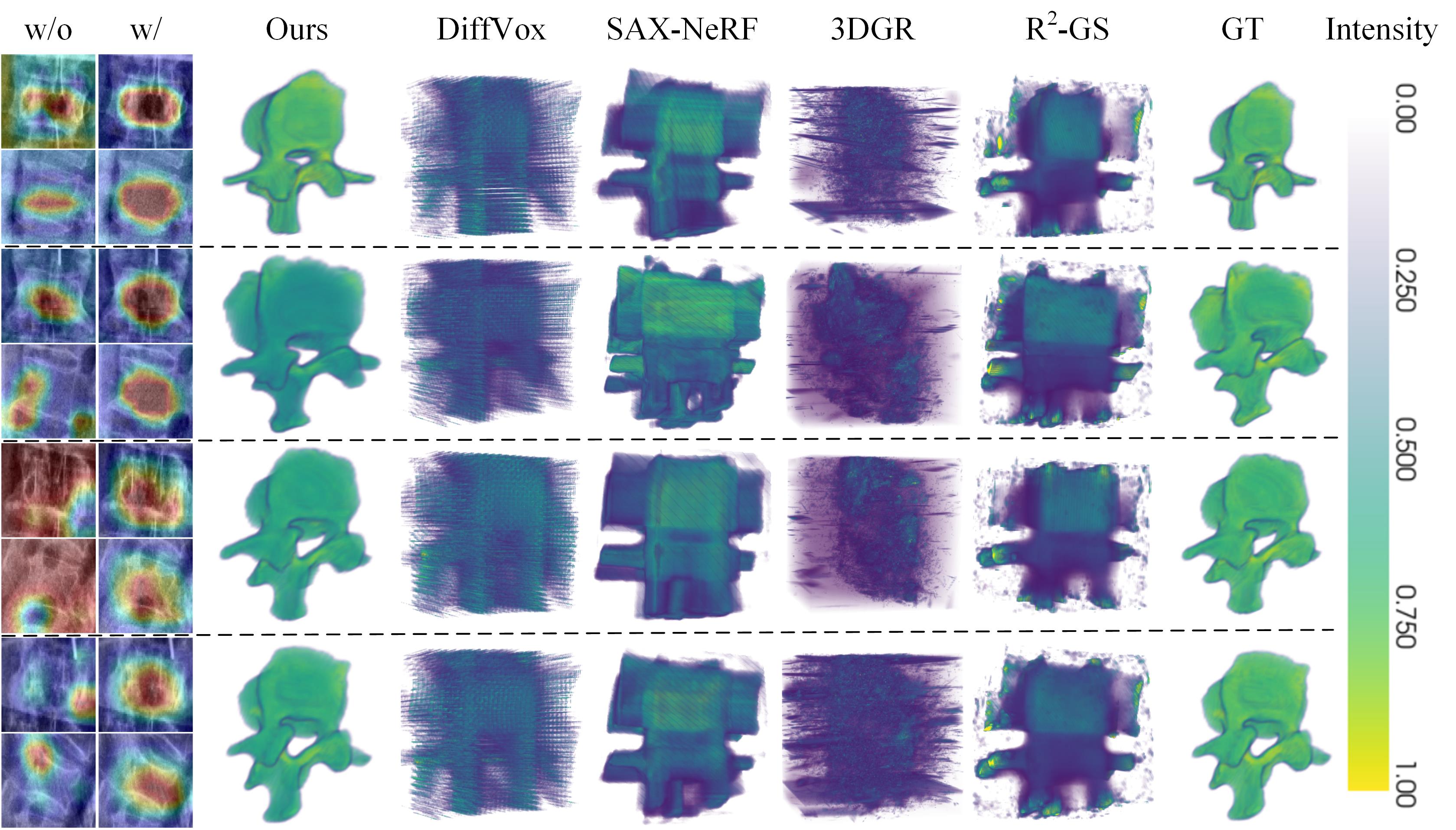}
\caption{Qualitative comparison with the different reconstruction methods. For each case separated by a dashed line, column 1 displays attention maps from RecM without CAL, while column 2 shows those with CAL, with the color bar indicating the voxel intensity distribution.} \label{recon_vis}
\end{figure}

\begin{figure}[ht]
\includegraphics[width=\textwidth]{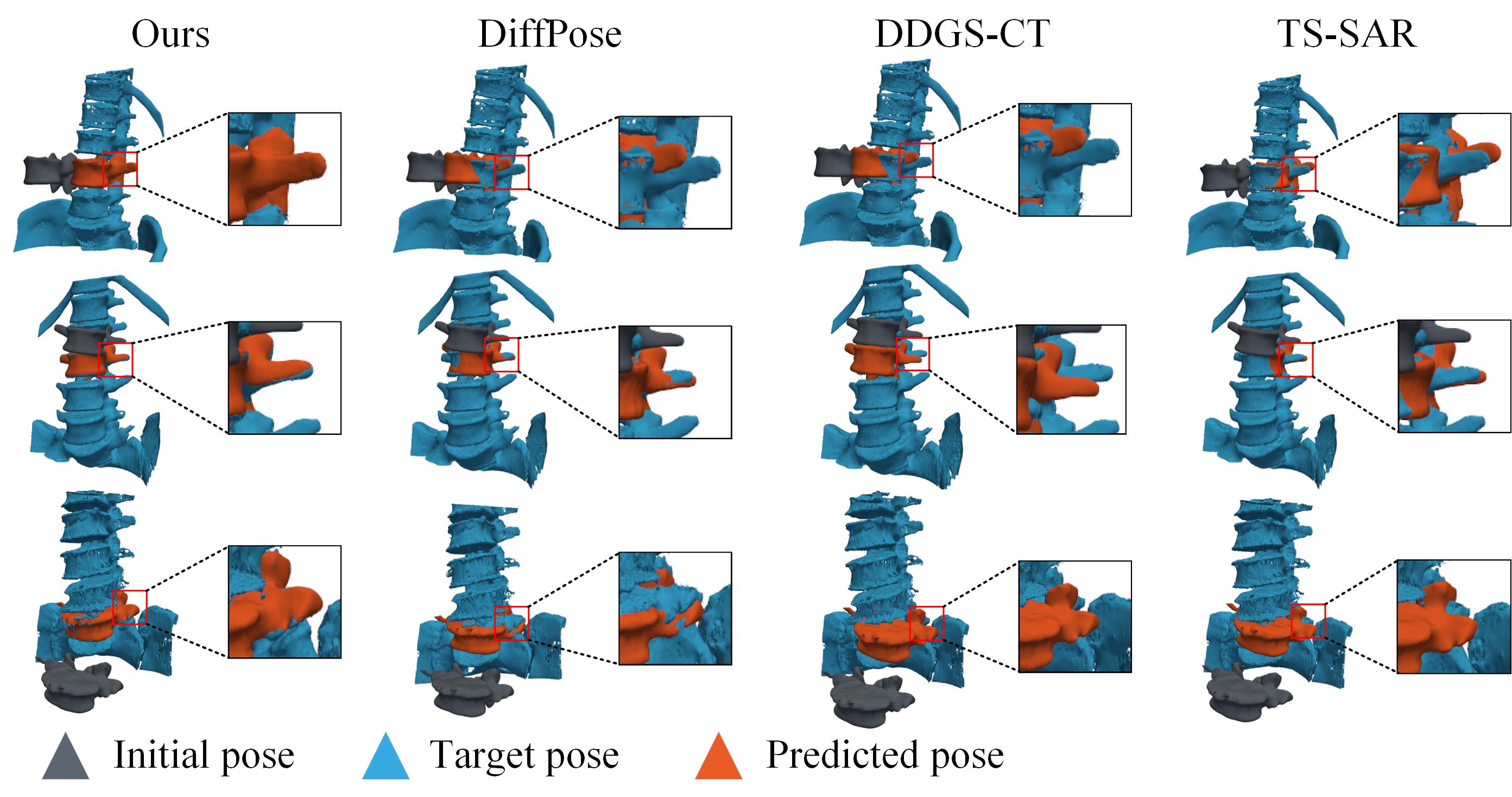}
\caption{Qualitative comparison with different registration methods. Each row represents an individual case.} \label{reg_vis}
\end{figure}

We conducted experiments on 3D reconstruction and registration, as detailed in Table~\ref{Compared}, with results marked by the dotted line. For reconstruction, we compared classical (DiffVox), NeRF-based (SAX-NeRF~\cite{cai2024structure}), and 3DGS-based methods (3DGR~\cite{fu20243dgr}, R$^2$-GS~\cite{zha2024r}), using $\mathcal{L}_{rec}$ from Eq. (\ref{eq:cal_recon}). RadGS-Reg achieves efficient, optimization-free reconstruction by encoding implicit vertebral \ac{CT} shape priors into an encoder, enabling inference based only on biplanar X-rays. For registration, we included learning-based (DiffPose~\cite{gopalakrishnan2024intraoperative}), 3DGS-based (DDGS-CT~\cite{gao2024ddgs}), and Lift3D-based (TS-SAR~\cite{zhao2024automatic}) methods. RadGS-Reg outperformed the others in both reconstruction and registration. As shown in Fig. \ref{recon_vis}, the shape and voxel intensity distribution of the vertebrae reconstructed by our method are more similar to the \ac{GT} volume. In Fig. \ref{reg_vis}, our approach yielded the most precise registration, with the preoperative vertebral \ac{CT} registering accurately to the intraoperative position. The intraoperative volume reconstructed from two X-rays elevated the 2D/3D registration to 3D space, and the registration facilitated by RecM effectively bridged the domain gap inherent in the "render and compare" approaches~\cite{gopalakrishnan2024intraoperative,gao2024ddgs}.

% In terms of reconstruction, a quantitative comparison for dual-view vertebra reconstruction is conducted among DiffVox, SAX-NeRF, 3DGR, r$^2$-GS, and ours. For registration, a quantitative comparison for vertebra registration is performed among DiffPose, DDGS-CT, TS-SAR, and ours (mean±std, best results in bold).

\begin{table}[ht]
\caption{Ablation study of our method. P$_1$, P$_2$, and P$_{Full}$ represent the first stage, first and second stage, and entire three stage pre-training, respectively.}
 \label{Ablation-exp}
 \centering
 \begin{tabular}{ll|cc|ccc}
    \hline
    \multirow{2}{*}{No.} & \multirow{2}{*}{Method} & \multicolumn{2}{c|}{Reconstruction} & \multicolumn{3}{c}{Registration} \\\cline{3-7}
 & & SSIM(\%) $\uparrow$ & PSNR(dB) $\uparrow$ & \ac{mTRE}(mm)$\downarrow$ & \ac{CR}(mm)$\uparrow$ & \ac{SR}(\%)$\uparrow$ \\
\hline
       (1) &RecM+P$_1$ &20.56±1.31 &14.87±0.31 &- &- &- \\
       (2) &RecM+CAL+P$_1$ &22.37±1.64 &15.45±0.47 &- &- &-\\

       (3) &RegM+P$_1$ &- &- &4.70±8.90 &0-5 &6.67  \\
       (4) &RadGS-Reg+P$_1$ &32.78±0.45 &18.99±0.17 &4.03±3.42 &5-10 &30.00 \\
       
     (5) &RadGS-Reg+P$_2$ &86.42±0.97 &27.29±0.30 &1.36±1.19  &15-20  &80.00 \\
     (6) &RadGS-Reg+P$_{Full}$ &\textbf{94.51±0.39} &\textbf{28.80±0.33} &\textbf{1.14±1.01} &\textbf{20-25} &\textbf{93.33}   \\
      (7)  &Dense-backbone &93.77±0.32 &28.27±0.26 &1.28±0.97 &20-25 &90.00   \\
      (8)  &ViT-backbone &94.08±1.09 &28.20±0.79 &1.27±0.95 &20-25 &93.33   \\

\hline
    \end{tabular}
\end{table}

\subsubsection{Ablation Study}
Ablation studies were conducted to assess the efficacy of the modules within RadGS-Reg, encompassing the CAL module, synergistic training, three stage pre-training, and model-agnostic characteristics, as delineated in Table \ref{Ablation-exp}. Results from experiments (1) and (2) validated that the CAL module enhanced the accuracy of vertebral reconstruction. Analysis of X-ray attention maps in Fig. \ref{recon_vis} indicated that the CAL module directed the network's focus towards the vertebral regions within the X-rays. A comparative analysis of experiment results (3) and (4) revealed that synergistic training substantially improved both registration and reconstruction performance. In this context, RadGS-Reg epitomized the synergistic training between RecM and RegM. Findings from experiments (4)-(6) demonstrated that exclusive training of RadGS-Reg on simulated data yielded suboptimal performance on the in-house dataset. Incrementally incorporating real X-rays and patient-specific DRRs, which provided shape priors, enabled RadGS-Reg to realize optimal performance across both tasks. Furthermore, a series of experiments employing various backbones (e.g., ResNet-50~\cite{he2016deep}, DenseNet-121~\cite{huang2017densely}, and ViT-12~\cite{dosovitskiy2020image}) in the reconstruction and registration modules of RadGS-Reg were conducted. The backbone network was uniformly configured as ResNet-50 for experiments (1)-(6). Findings from experiments (6)-(8) demonstrated that the proposed synergistic training possessed model-agnostic properties, consistently achieving favorable outcomes across both tasks, irrespective of the backbone employed.

\section{Conclusion}
We present RadGS-Reg, a unified framework for accurate and efficient vertebral CT/X-ray registration by synergizing 3D \ac{RadGS} reconstruction and 3D/3D registration. Key innovations include the integration of \ac{CAL} to mitigate vertebral overlap interference, a module-aware synergistic training strategy, and a three stage data-aware pre-training strategy. RadGS-Reg significantly outperforms existing methods in reconstruction quality (94.51\% SSIM) and registration accuracy (1.14 mm mTRE). Its model-agnostic design ensures flexibility, while its 0.82s runtime meets real-time clinical demands.  Future work will extend the applicability of RadGS-Reg by: 1) incorporating deformation models for complex non-rigid regions involving soft tissues and vasculature, and 2) implementing post-reconstruction Gaussian segmentation to streamline the current preprocessing procedures.

% Future work will extend RadGS-Reg to non-rigid registration and multi-anatomy applications, enhancing its utility in broader surgical navigation contexts.

\subsubsection{Acknowledgments.} This work was supported partly by Jiangsu Province Key Research Program-Social Development-Clinical Frontier Technologies (BE2022718).
\subsubsection{Disclosure of Interests.} The authors declare that they have no known competing financial interests or personal relationships that could have appeared to influence the work reported in this article.

\bibliographystyle{splncs04}
\bibliography{paper-2299}




% LNCS Homepage, \url{http://www.springer.com/lncs}, last accessed 2023/10/25
% \end{thebibliography}
\end{document}